\documentclass{article}
\usepackage{PRIMEarxiv}

\usepackage[utf8]{inputenc} 
\usepackage[T1]{fontenc}    
\usepackage{hyperref}       
\usepackage{url}            
\usepackage{booktabs}       
\usepackage{amsfonts}       
\usepackage{nicefrac}       
\usepackage{microtype}      
\usepackage{lipsum}
\usepackage{fancyhdr}       
\usepackage{graphicx}       
\graphicspath{{media/}}     
\usepackage{xcolor}
\usepackage{float}
\usepackage{natbib}
\definecolor{Gray}{gray}{0.6}
\usepackage{xcolor}
\usepackage{subcaption}
\title{Optimizing Transformer-based Machine Translation Model for Single GPU Training: A Hyperparameter Ablation Study}

\author{Luv Verma\thanks{\textit{corresponding author}} \\
    Khoury College of Computer Sciences\\
    Northeastern University\\
    Boston, MA 02115 \\
    \texttt{verma.lu@northeastern.edu} \and 
    \textbf{Ketaki N. Kolhatkar}\\
    Khoury College of Computer Sciences\\
    Northeastern University\\
    Boston, MA 02115 \\
    \texttt{kolhatkar.k@northeastern.edu}
}

\begin{document}
\maketitle

\begin{abstract}

In machine translation tasks, the relationship between model complexity and performance is often presumed to be linear, driving an increase in the number of parameters and consequent demands for computational resources like multiple GPUs. To explore this assumption, this study systematically investigates the effects of hyperparameters through ablation on a sequence-to-sequence machine translation pipeline, utilizing a single NVIDIA A100 GPU. Contrary to expectations, our experiments reveal that combinations with the most parameters were not necessarily the most effective. This unexpected insight prompted a careful reduction in parameter sizes, uncovering "sweet spots" that enable training sophisticated models on a single GPU without compromising translation quality. The findings demonstrate an intricate relationship between hyperparameter selection, model size, and computational resource needs. The insights from this study contribute to the ongoing efforts to make machine translation more accessible and cost-effective, emphasizing the importance of precise hyperparameter tuning over mere scaling.

\end{abstract}


\section{Introduction}
The advent of technology has propelled humankind into a new era where machine learning is progressing rapidly. With every passing day, models in fields such as data engineering, finance, computer vision, and natural language processing are growing both in size and capability. However, a pivotal question arises: Is bigger always better? In our current studies, we approached this question from two perspectives. First, we delved into the intricacies of multiprocessing within Python using CUDA \cite{nvidia2023course}. Second, we selected a complex machine translation task in the domain of natural language processing, requiring a complete sequence-to-sequence pipeline \cite{vaswani2017attention}, to examine the relationship between model size and performance.\\
Traditionally, the enhancement of model performance has been synonymous with an increase in parameters, layers, and complexity. Yet, we must confront the reality that the relationship between size and efficacy is not necessarily linear, especially when considering the need for efficiency, stability, and resource-conscious design. Our journey began with an ambitious plan to utilize both model parallelism and data parallelism in PyTorch, with a starting model comprising 500 million parameters. However, constraints such as the availability of multi-GPUs and a limited compute window of 8 hours on a single A100 GPU \cite{nvidia2020a100} steered us toward a different path. Instead of continually expanding the model, we embarked on a gritty exploration to reduce the model's parameters.\\
We conducted ablation studies, manipulating variables like model size, number of heads, layers, and dropouts, with a computation limit on an A100 single GPU. By capping the number of epochs to 100, we investigated the effects of varying parameter sizes on the accuracies of machine translation models. Our findings culminated in the fascinating observation that an increase in parameters does not always yield a better model, a notion aligned with recent research conducted in Chinchilla paper \cite{hoffmann2022training}. This study underscores the idea that training a robust model does not necessarily require a multi-GPU setup. Before scaling up hardware, it is vital to meticulously analyze how the model is performing within existing resources and understand how the model's size correlates with accuracy.

\section{Experimental Details}
In this section, we delve into the ablation studies conducted to unravel the intricate relationship between the model's size, number of parameters, running time, and accuracy. The study was guided by a key question: Does continuously increasing the model size invariably lead to an increase in accuracy? Through meticulous experimentation and hyperparameter tuning, this investigation aimed to answer this question. As an example, a machine translation task (english to spanish) was taken, as it utilizes complete encoder and decoder pipeline.

\subsection{Data Preparation and Preprocessing}
The experimental study focused on the machine translation task from English to Spanish. The dataset comprising 1 lakh translations was utilized, sourced from Kaggle (can be directly downloaded from tensorflow reference \cite{tensorflow2023spanish-english}). The chosen size of the problem was designed to experiment with the relationship between the model's size, number of parameters, and its performance. The code is available on GitHub \cite{luvverma2011}

\subsection{Model Architecture}
The Transformer model architecture was utilized in this study, renowned for its parallelization and the ability to capture long-range dependencies in the sequence data. The model consists of an encoder and decoder, each with multiple layers of attention and feed-forward neural networks. The following hyperparameters were explored:

\begin{itemize}
\item \textbf{Model Size}: Ranging from 16 to 512, reflecting the embedding dimensions.
\item \textbf{Number of Heads}: Values varied from 4 to 16, influencing the model's attention mechanism.
\item \textbf{Number of Layers}: Tested with 2 to 16 layers, impacting the depth of the network.
\item \textbf{Dropout Sizes}: Ranging from 0.1 to 0.5, to prevent overfitting.
\item \textbf{Number of Epochs}: Since, this study was about evaluating the effect of increase in number of parameters on model accuracy, the most of the cases were limited to 100 epochs. A few best cases were run till 400 epochs.

\end{itemize}

\subsection{Training Configuration}
Training was conducted on a single NVIDIA A100 GPU. The model's loss was computed using the cross-entropy loss function for the machine translation task. A training-validation split of 70-30 \% was employed.

\section{Results and Discussions}
\label{sec:headings}
\subsection{Results with model size of 512}
\begin{figure}[ht]
\centering
\includegraphics[width=150mm]{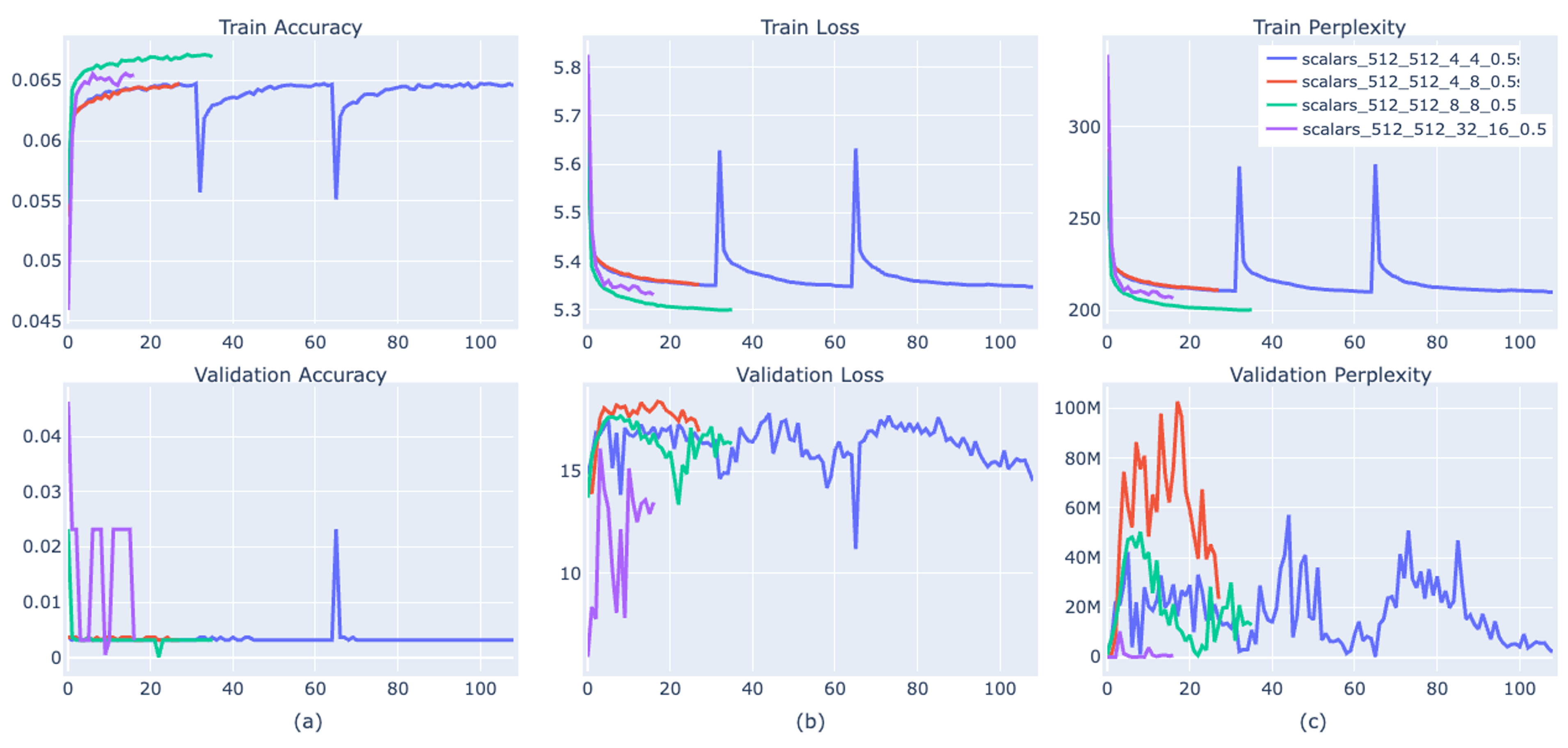}
\caption{Experiments: Maximum length = 512, Model Size = 512, number of heads = 4, 8, 32, number of layers = 4, 8, 16, and dropout sizes = 0.5 (combinations shown in legends). Ran for 400 epochs approx. (x-axis). \textbf{(a)} Train and Validation Accuracy: Comparison of training and validation accuracy across different configurations. \textbf{(b)} Train and Validation Loss: Plot of training and validation loss for different models. \textbf{(c)} Train and Validation Perplexity: Visualization of training and validation perplexity under various hyperparameter settings.}
\label{fig2}
\end{figure}

\begin{figure}[ht]
\centering
\includegraphics[width=150mm]{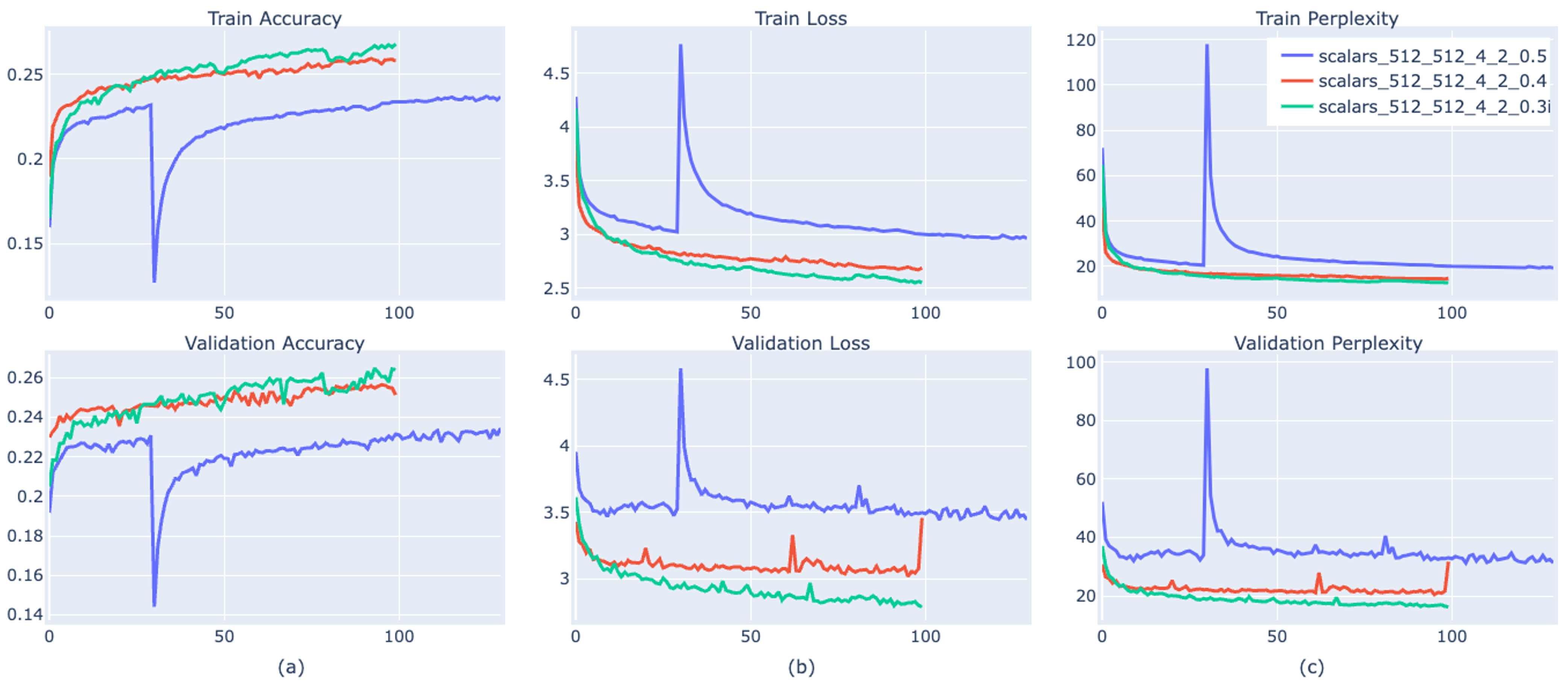}
\caption{Experiments: Maximum length = 512, Model Size = 512, number of heads = 4, number of layers = 2, and dropout sizes = 0.3, 0.4, 0.5, (combinations shown in legends). Ran for 100 epochs (x-axis). \textbf{(a)} Train and Validation Accuracy. \textbf{(b)} Train and Validation Loss. \textbf{(c)} Train and Validation Perplexity.}
\label{fig3}
\end{figure}

The results presented in Figure \ref{fig2} reveal a disconcerting instability, particularly apparent when the model size is at its peak value of 512, with both the number of heads and layers set at 4. The instability in learning is suspected to be linked to a dropout value of 0.5, a feature typically employed to prevent overfitting through the introduction of noise. However, in this context, the chosen dropout value appears excessively high, inhibiting convergence and potentially suppressing essential features. This is exacerbated by the validation perplexity reaching into the millions, a stark and troubling indicator of the model's learning instability. With a reduction in the dropout value and the number of layers to 2, as depicted in Figure \ref{fig3}, the model begins to exhibit signs of learning, despite a persistently rugged learning surface. This intriguing pattern suggests an intricate interplay between the model's complexity (and thus parameter count) and the effectiveness of learning with a reduced dropout value.
Figure \ref{fig3} offers a closer examination of this phenomenon. Although learning begins to occur, convergence is not achieved. Validation losses and perplexity remain elevated even when the dropout is decreased to 0.3. An anomalous 'kink' in validation perplexity observed around the 100th epoch for a dropout of 0.4 (as illustrated in Figure \ref{fig3}(b)) may hint at the model's need for additional epochs to stabilize. This interpretation, though compelling, requires more rigorous testing or analysis to be validated.
\begin{figure}[ht]
\centering
\includegraphics[width=150mm]{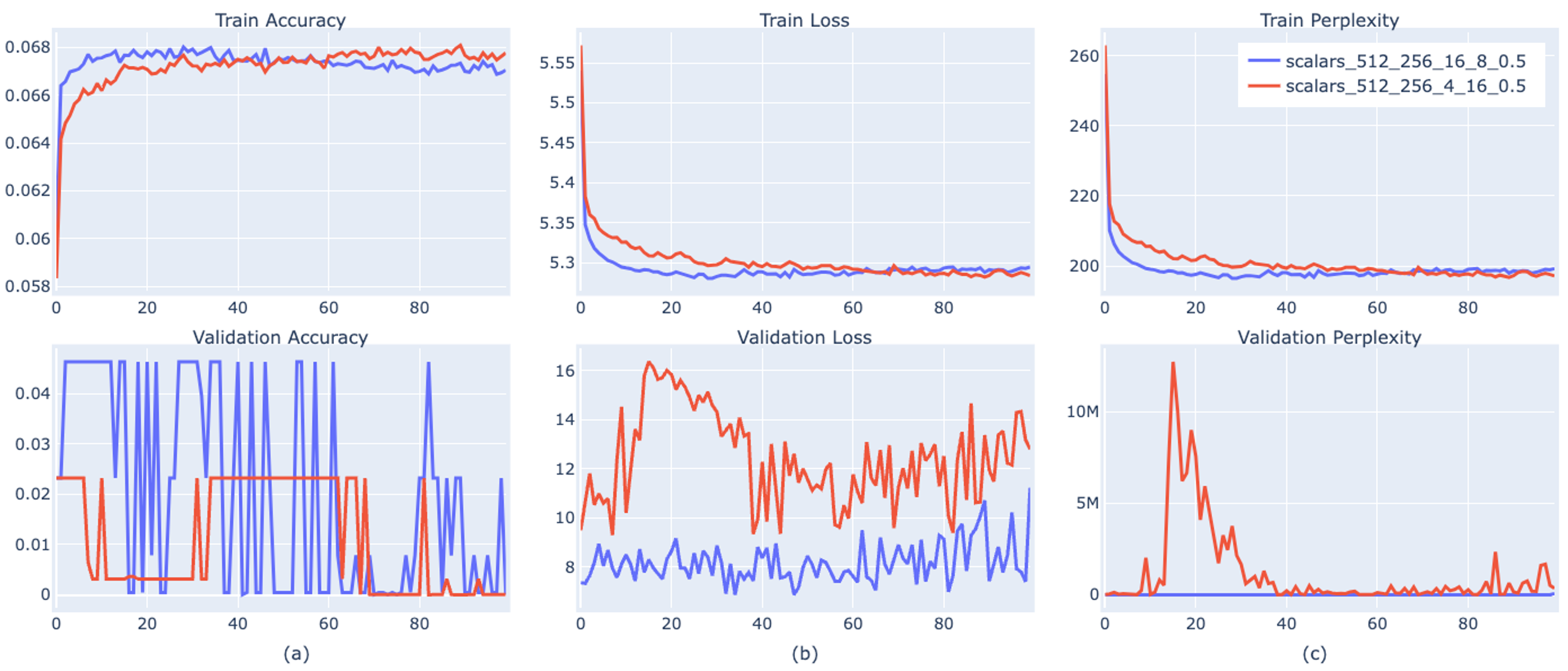}
\caption{Experiments: Maximum length = 512, Model Size = 256, number of heads = 4, 16, number of layers = 8,16, and dropout sizes = 0.5 (combinations shown in legends). Ran for 100 epochs (x-axis). \textbf{(a)} Train and Validation Accuracy. \textbf{(b)} Train and Validation Loss. \textbf{(c)} Train and Validation Perplexity.}
\label{fig4}
\end{figure}
\begin{figure}[ht]
\centering
\includegraphics[width=150mm]{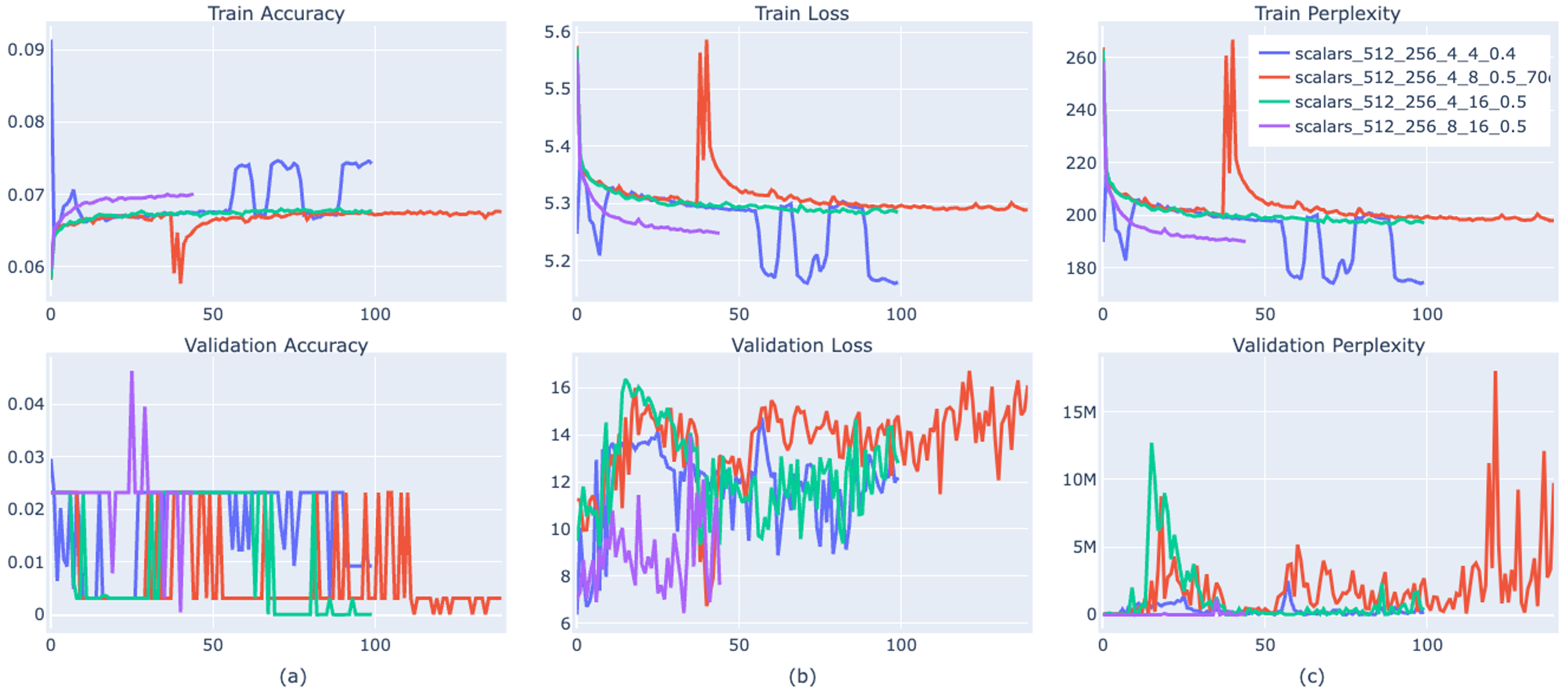}
\caption{Experiments: Maximum length = 512, Model Size = 256, number of heads = 4, 8, number of layers = 4, 8, 16, and dropout sizes = 0.4, 0.5 (combinations shown in legends). Ran for 100 epochs (x-axis). \textbf{(a)} Train and Validation Accuracy. \textbf{(b)} Train and Validation Loss. \textbf{(c)} Train and Validation Perplexity.}
\label{fig5}
\end{figure}
\begin{figure}[ht]
\centering
\includegraphics[width=150mm]{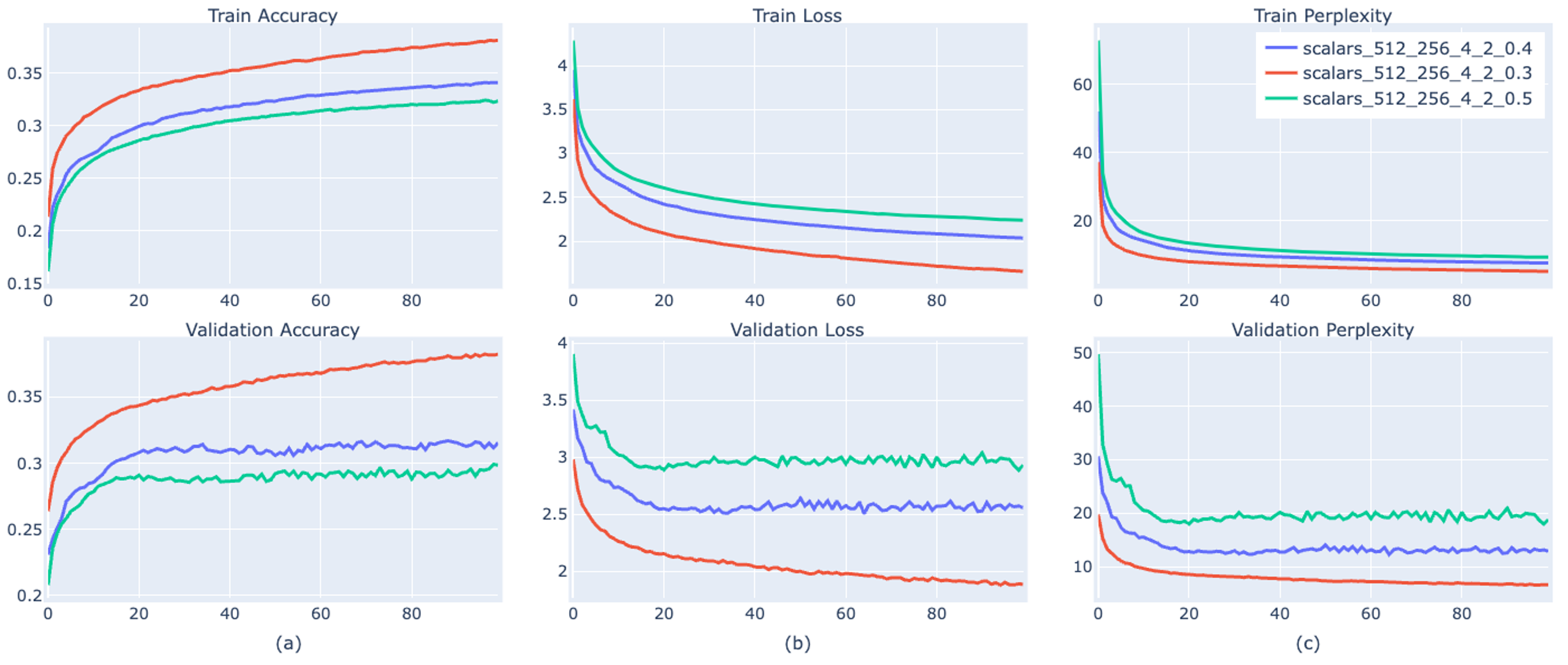}
\caption{Experiments: Maximum length = 512, Model Size = 256, number of heads = 4, number of layers = 2, and dropout sizes = 0.3, 0.4, 0.5 (combinations shown in legends). Ran for 100 epochs (x-axis). \textbf{(a)} Train and Validation Accuracy. \textbf{(b)} Train and Validation Loss. \textbf{(c)} Train and Validation Perplexity.}
\label{fig6}
\end{figure}
\begin{figure}[ht]
\centering
\includegraphics[width=150mm]{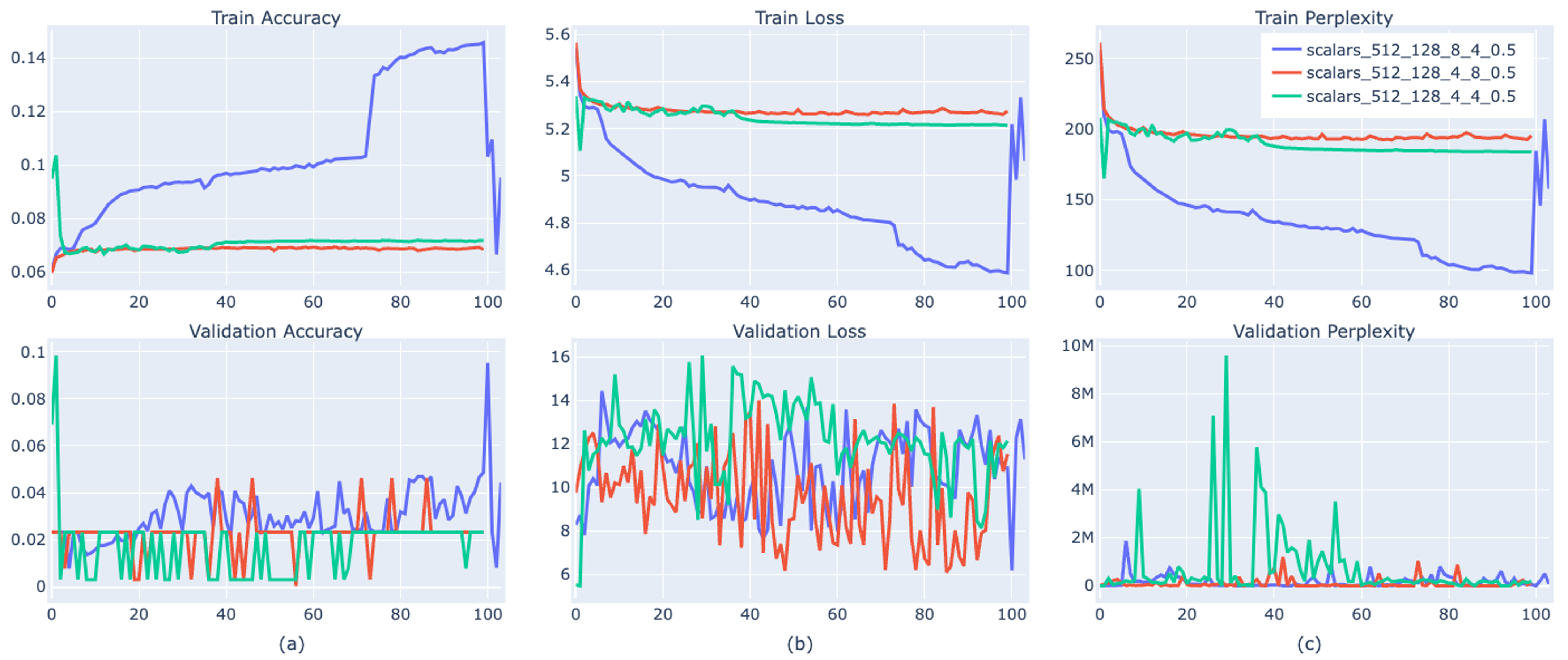}
\caption{Experiments: Maximum length = 512, Model Size = 128, number of heads = 4, 8, number of layers = 4, 8, and dropout sizes = 0.5 (combinations shown in legends). Ran for 100 epochs (x-axis). \textbf{(a)} Train and Validation Accuracy. \textbf{(b)} Train and Validation Loss. \textbf{(c)} Train and Validation Perplexity.}
\label{fig7}
\end{figure}
\begin{figure}[ht]
\centering
\includegraphics[width=150mm]{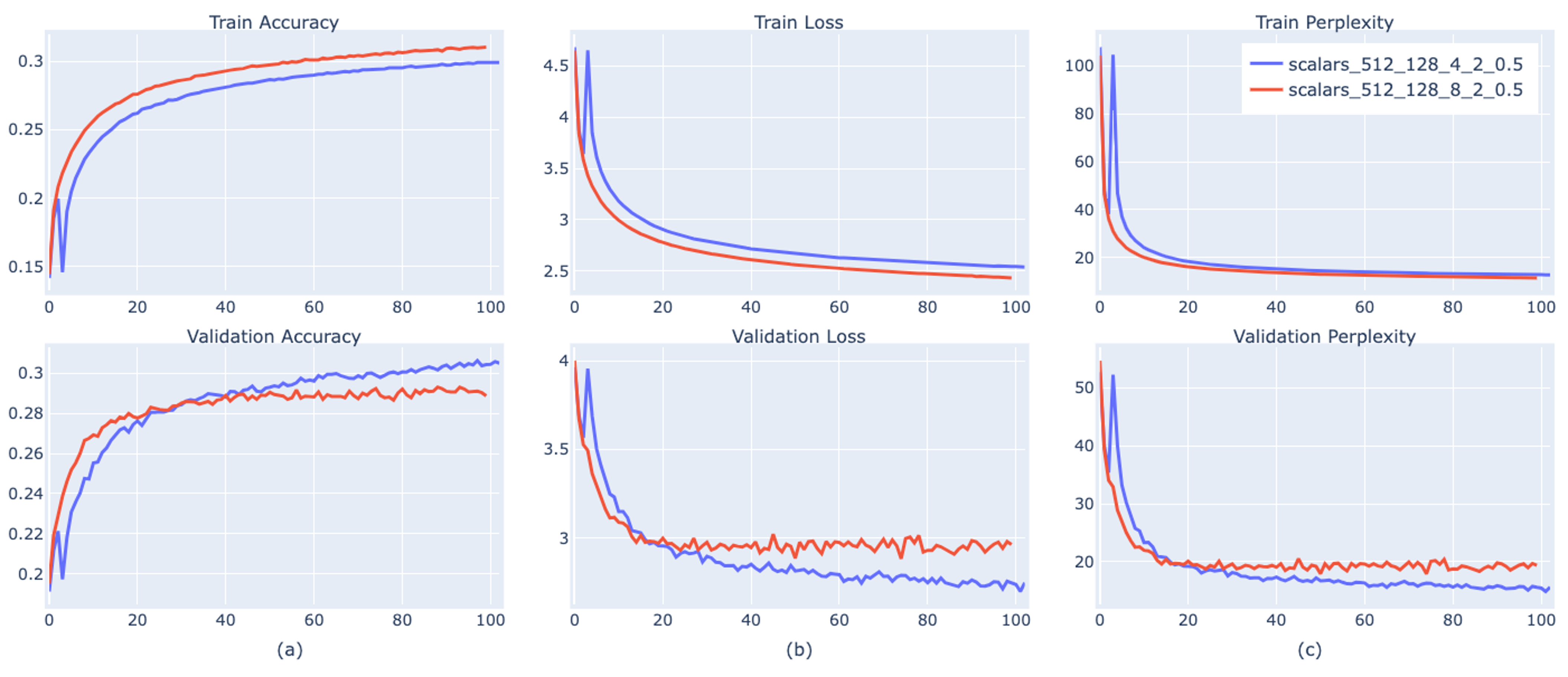}
\caption{Experiments: Maximum length = 512, Model Size = 128, number of heads = 4, 8, number of layers = 2, and dropout sizes = 0.5 (combinations shown in legends). Ran for 100 epochs (x-axis). \textbf{(a)} Train and Validation Accuracy. \textbf{(b)} Train and Validation Loss. \textbf{(c)} Train and Validation Perplexity.}
\label{fig8}
\end{figure}
\begin{figure}[ht]
\centering
\includegraphics[width=150mm]{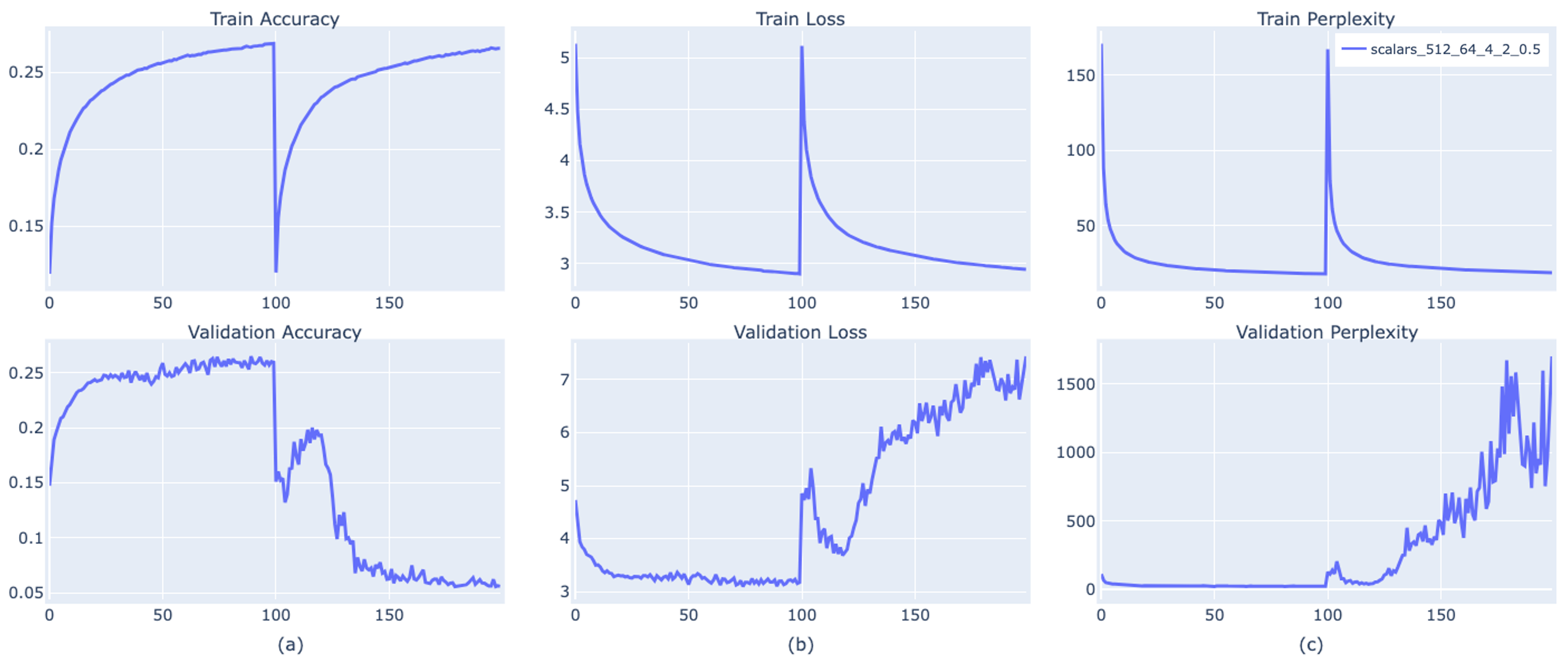}
\caption{Experiments: Maximum length = 512, Model Size = 64, number of heads = 4, number of layers = 2, and dropout sizes = 0.5 (combinations shown in legends). Ran for 100 epochs (x-axis). \textbf{(a)} Train and Validation Accuracy. \textbf{(b)} Train and Validation Loss. \textbf{(c)} Train and Validation Perplexity.}
\label{fig9}
\end{figure}
\begin{figure}[ht]
\centering
\includegraphics[width=150mm]{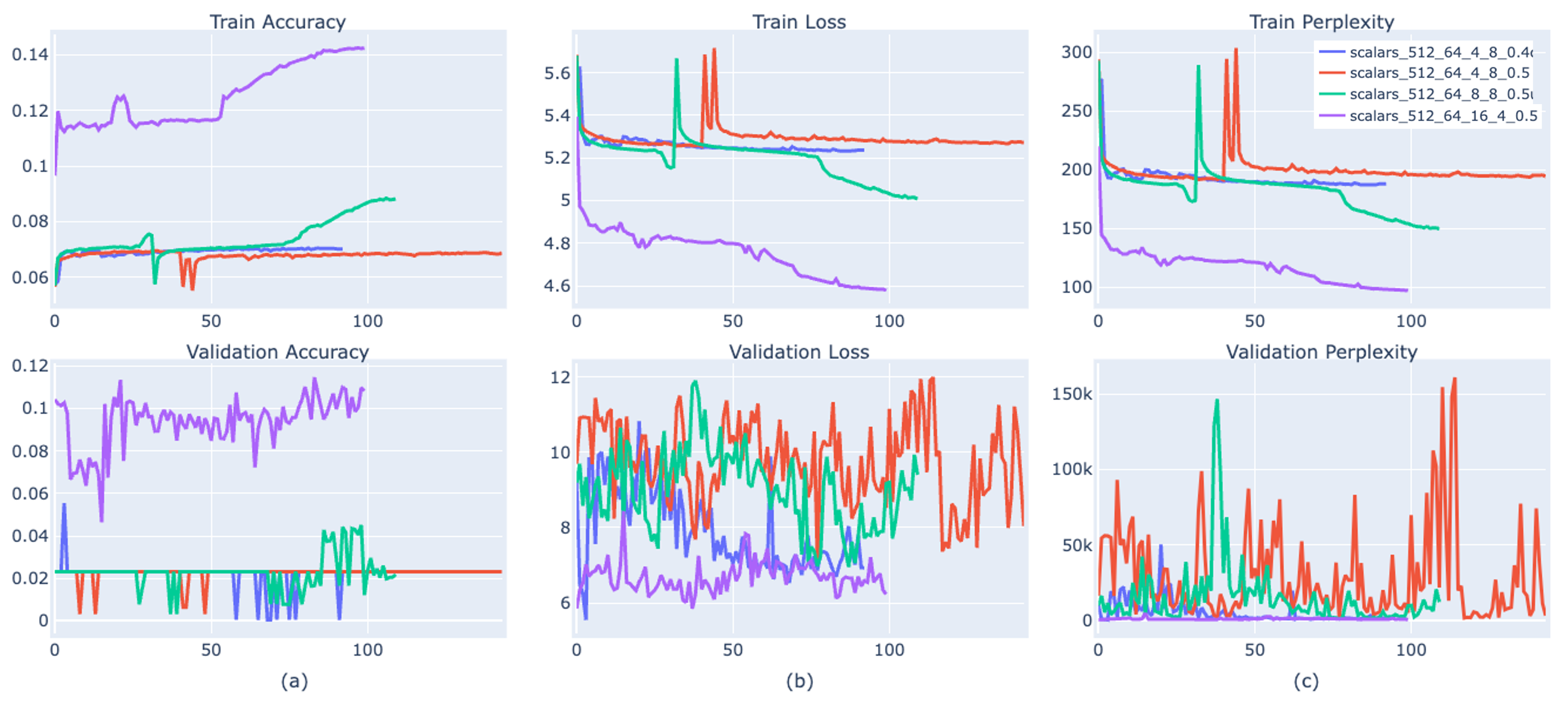}
\caption{Experiments: Maximum length = 512, Model Size = 64, number of heads = 4, 8, 16, number of layers = 4, 8, and dropout sizes = 0.5 (combinations shown in legends). Ran for 100 epochs (x-axis). \textbf{(a)} Train and Validation Accuracy. \textbf{(b)} Train and Validation Loss. \textbf{(c)} Train and Validation Perplexity.}
\label{fig10} 
\end{figure}
\begin{figure}[ht]
\centering
\includegraphics[width=150mm]{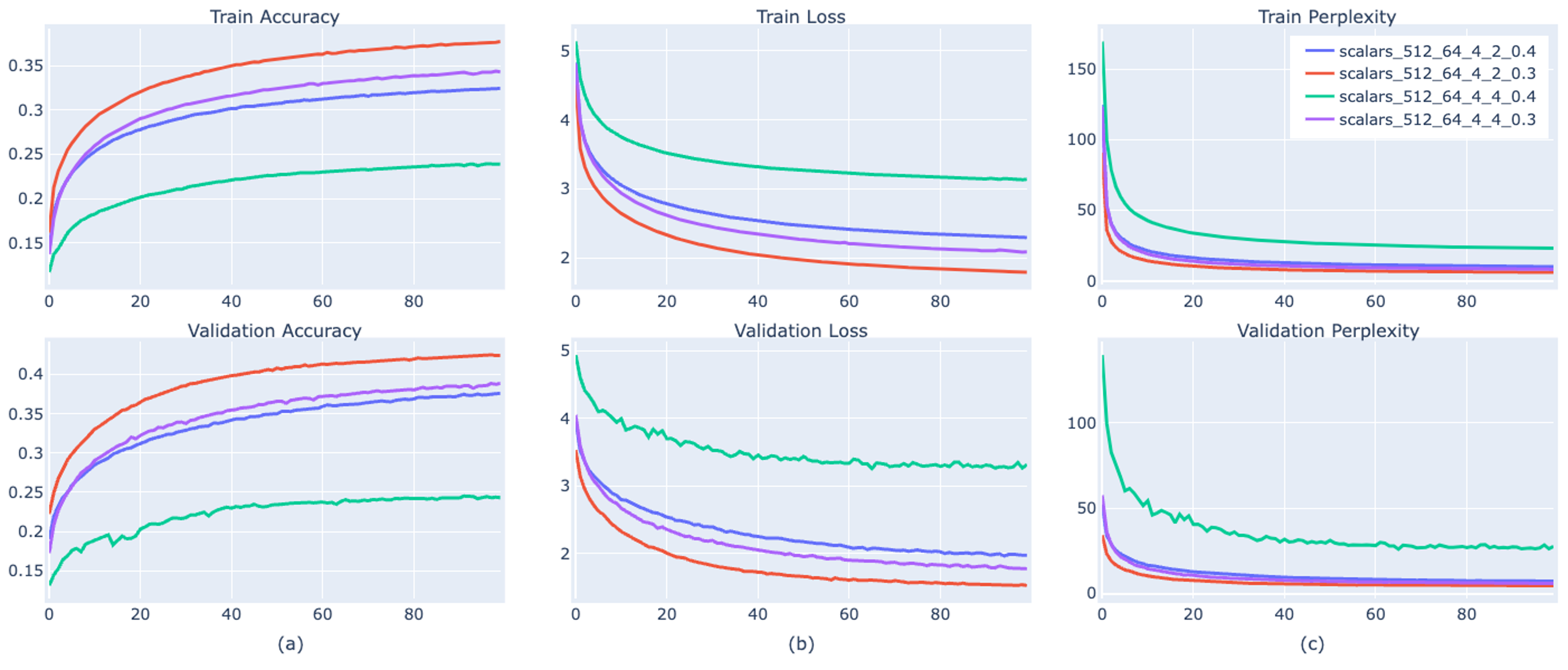}
\caption{Experiments: Maximum length = 512, Model Size = 64, number of heads = 4, number of layers = 2, 4, and dropout sizes = 0.3, 0.4 (combinations shown in legends). Ran for 100 epochs (x-axis). \textbf{(a)} Train and Validation Accuracy. \textbf{(b)} Train and Validation Loss. \textbf{(c)} Train and Validation Perplexity.}
\label{fig11} 
\end{figure}
\subsection{Results with model size of 256}
Figure \ref{fig4} presents a further reduction in model size to 256, exploring two specific combinations of heads and layers. For a configuration of 16 heads and 8 layers, the learning is minimal, denoted by a train accuracy improvement on the order of 1e-3. Conversely, the model demonstrates overfitting and becomes unstable, the ruggedness augmenting with increasing epochs. When heads are reduced to 4 and layers increased to 16, learning fails to occur, and perplexity surpasses 10 million for roughly 20 epochs. The enormity of perplexity suggests that more epochs may not salvage this configuration. In another set of experiments for a model size of 256 (Figure \ref{fig5}), the combination of 4 heads and 8 layers (with a dropout of 0.5) led to the model becoming unstable, starting to overfit, and displaying erratic perplexity. Other experimental combinations, such as 4 heads with 16 layers and a dropout of 0.5, showed overfitting; the setup with 8 heads, 16 layers, and a dropout of 0.5 was halted midway due to the onset of enormous perplexity scores and evident overfitting
\begin{table}
\centering
\caption{An analysis of the impact of different hyperparameters on machine translation model performance, along with the corresponding parameter count for each configuration (\textbf{for 100 epochs})}
\label{table1}
\begin{tabular}{lllllllll}
\toprule
  model & number of & number of & dropout & Time (min) & Train & Validation & Validation & Parameter \\
  size & heads & layers & & & Loss & Loss & Perplexity & Count (million) \\
\midrule
  16 & 4 & 2 & 0.1 & 71.0 & 2.7389 & 2.3684 & 10.6806 & 3.20 \\
  16 & 4 & 4 & 0.1 & 93.0 & 2.9223 & 2.5788 & 13.1816 & 3.21 \\
  16 & 8 & 2 & 0.5 & 70.0 & 4.3441 & 4.477 & 87.9667 & 3.20 \\
  16 & 4 & 2 & 0.5 & 71.0 & 4.4678 & 6.1332 & 460.9081 & 3.20 \\
  32 & 4 & 4 & 0.1 & 98.0 & 1.8019 & 1.5429 & 4.6783 & 6.42 \\
  32 & 4 & 2 & 0.1 & 67.0 & 1.669 & 1.4727 & 4.361 & 6.36 \\
  32 & 4 & 2 & 0.5 & 67.0 & 3.7191 & 10.8591 & 5.2e+04 & 6.36 \\
  16 & 4 & 2 & 0.5 & 71.0 & 4.4678 & 6.1332 & 460.9081 & 3.20 \\
  32 & 8 & 2 & 0.5 & 68.0 & 3.6843 & 5.202 & 181.6349 & 6.36 \\
  64 & 4 & 8 & 0.5 & 244.0 & 5.2703 & 8.035 & 3.1e+03 & 13.48 \\
  64 & 8 & 8 & 0.5 & 176.0 & 5.0072 & 9.399 & 1.2e+04 & 13.48 \\
  64 & 16 & 4 & 0.5 & 98.0 & 4.5781 & 6.232 & 508.7538 & 13.01 \\
  64 & 4 & 2 & 0.4 & 71.0 & 2.2984 & 1.9782 & 7.23 & 12.78 \\
  64 & 4 & 2 & 0.3 & 71.0 & 1.7956 & 1.5268 & 4.6035 & 12.78 \\
  64 & 4 & 4 & 0.4 & 94.0 & 3.1409 & 3.3198 & 27.6538 & 13.01 \\
  64 & 4 & 4 & 0.3 & 92.0 & 2.0916 & 1.7769 & 5.9116 & 13.01 \\
  64 & 4 & 2 & 0.5 & 121.0 & 2.9436 & 7.4402 & 1.7e+03 & 12.78 \\
  128 & 4 & 2 & 0.1 & 79.0 & 0.5859 & 0.9217 & 2.5136 & 25.95 \\
  \textbf{128} & \textbf{4} & \textbf{4} & \textbf{0.1} & \textbf{107.0} & \textbf{0.6481} & \textbf{0.8993} & \textbf{2.4579} & \textbf{26.87} \\
  128 & 4 & 2 & 0.5 & 79.0 & 2.5408 & 2.748 & 15.6114 & 25.95 \\
  128 & 8 & 2 & 0.5 & 79.0 & 2.4338 & 2.9636 & 19.3666 & 25.95 \\
  128 & 8 & 4 & 0.5 & 110.0 & 5.062 & 11.2901 & 8.0e+04 & 26.87 \\
  128 & 4 & 8 & 0.5 & 178.0 & 5.2745 & 11.5297 & 1.0e+05 & 28.73 \\
  128 & 4 & 4 & 0.5 & 106.0 & 5.2141 & 12.1386 & 1.9e+05 & 26.87 \\
  256 & 4 & 2 & 0.1 & 94.0 & 0.3654 & 1.0448 & 2.8427 & 53.67 \\
  256 & 4 & 4 & 0.1 & 120.0 & 1.4242 & 1.9795 & 7.2392 & 57.36 \\
  256 & 4 & 2 & 0.4 & 94.0 & 2.036 & 2.5577 & 12.906 & 53.67 \\
  256 & 4 & 2 & 0.3 & 93.0 & 1.6566 & 1.8869 & 6.599 & 53.67 \\
  256 & 4 & 2 & 0.5 & 94.0 & 2.2395 & 2.9339 & 18.8001 & 53.67 \\
  256 & 16 & 8 & 0.5 & 209.0 & 5.2948 & 11.2313 & 7.5e+04 & 64.73 \\
  256 & 4 & 16 & 0.5 & 314.0 & 5.2841 & 12.7906 & 3.6e+05 & 79.47 \\
  256 & 4 & 16 & 0.5 & 314.0 & 5.2841 & 12.7906 & 3.6e+05 & 79.47 \\
  512 & 4 & 4 & 0.5 & 166.0 & 5.3473 & 14.5356 & 2.1e+06 & 129.34 \\
  512 & 4 & 2 & 0.5 & 229.0 & 2.9624 & 3.4459 & 31.371 & 114.62 \\
  512 & 4 & 2 & 0.4 & 135.0 & 2.6893 & 3.4579 & 31.7515 & 114.62 \\
  512 & 4 & 2 & 0.3 & 133.0 & 2.5512 & 2.7891 & 16.2657 & 114.62 \\
\bottomrule
\end{tabular}
\end{table}
\subsection{Results with the model size of 128}
We further dropped the model size to 128. In Figure \ref{fig7}, the only configuration that showed signs of learning was with 8 heads and 4 layers, combined with a dropout of 0.5. However, near the end of the 98th epoch, this setup suddenly became unstable, with training accuracy plunging to 0.06 from 0.14. This demonstrated that none of the three combinations above were effective.
The situation improved when the number of layers was reduced to 2 instead of 4 (Figure \ref{fig8}). This demonstrated that the model was learning with a lesser number of layers. From the validation accuracy and loss, it was observed that with a reduced number of heads in each layer (4 instead of 8), learning was more effective, and this progress continued up until 100 epochs.
\subsection{Results with the model size of 64}
In Figure \ref{fig9}, we further reduced the model size to 64 (from 128), keeping the number of heads, layers, and the dropout value the same as in Figure \ref{fig8}. We observed a peculiar trend where the model learned for 100 epochs, then regressed, and eventually started overfitting, as training loss reduced while validation loss increased. This curious trend prompted us to attempt training for more epochs in select cases.
In Figure \ref{fig10}, with the model size kept at 64 and the number of heads increased to 4, 8, and 16 (and layers increased to 8 with a dropout of 0.5), learning became unstable. Interestingly, when the number of heads was 16 and layers were 4, the model performed better than other cases in Figure \ref{fig10}. This indicates that there might be a scope for improvement if trained for more epochs. From Figure \ref{fig11}, for a model size of 64, with 4 heads and either 2 or 4 layers (and dropouts of 0.3 and 0.4), the model started learning without any signs of overfitting. It is quite evident from this figure that increasing the number of encoder-decoder layers reduced the learning rate, even with a higher dropout value of 0.4.
\subsection{Results with the model size of 32}
\begin{figure}[ht]
\centering
\includegraphics[width=150mm]{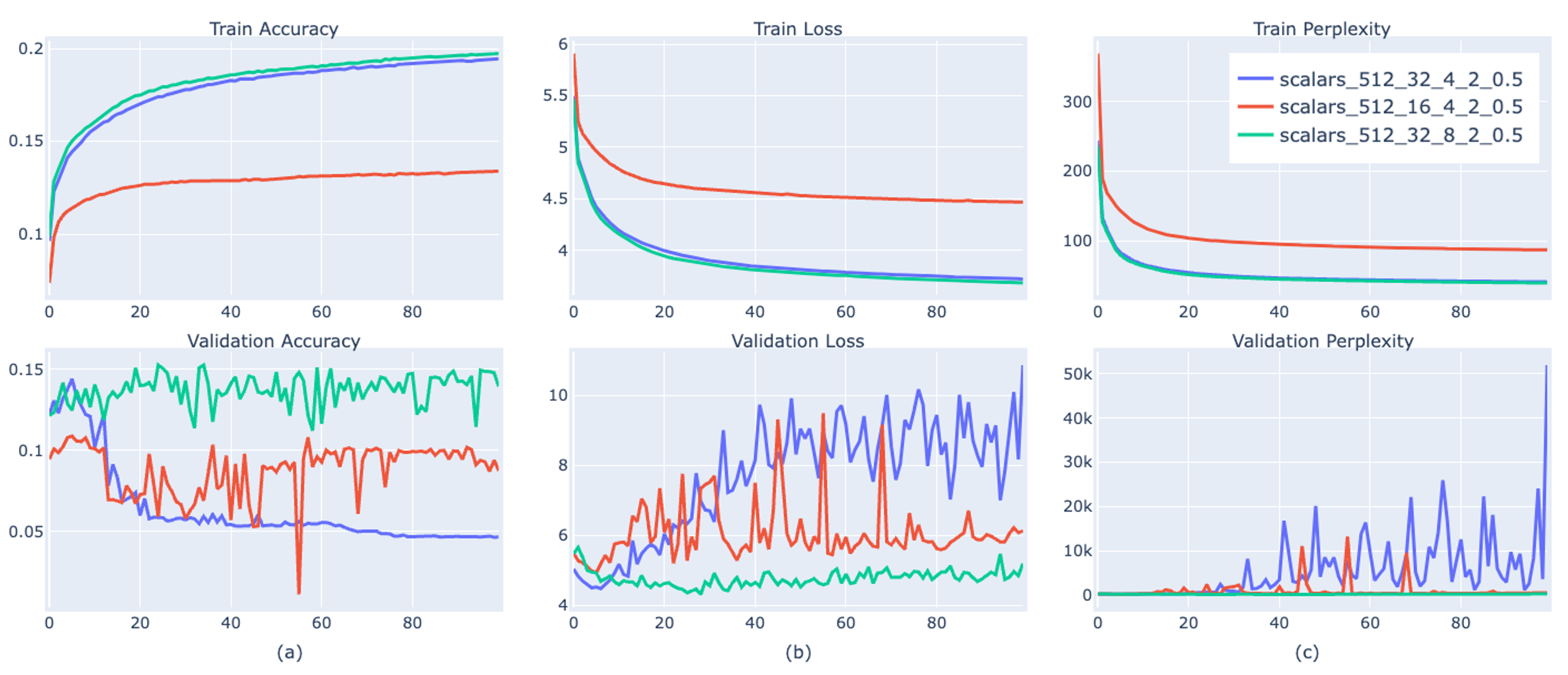}
\caption{Experiments: Maximum length = 512, Model Size = 32, 16, number of heads = 4, 8, number of layers = 2, and dropout sizes = 0.5 (combinations shown in legends). Ran for 100 epochs (x-axis). \textbf{(a)} Train and Validation Accuracy. \textbf{(b)} Train and Validation Loss. \textbf{(c)} Train and Validation Perplexity.}
\label{fig12}
\end{figure}

\begin{figure}[ht]
\centering
\includegraphics[width=150mm]{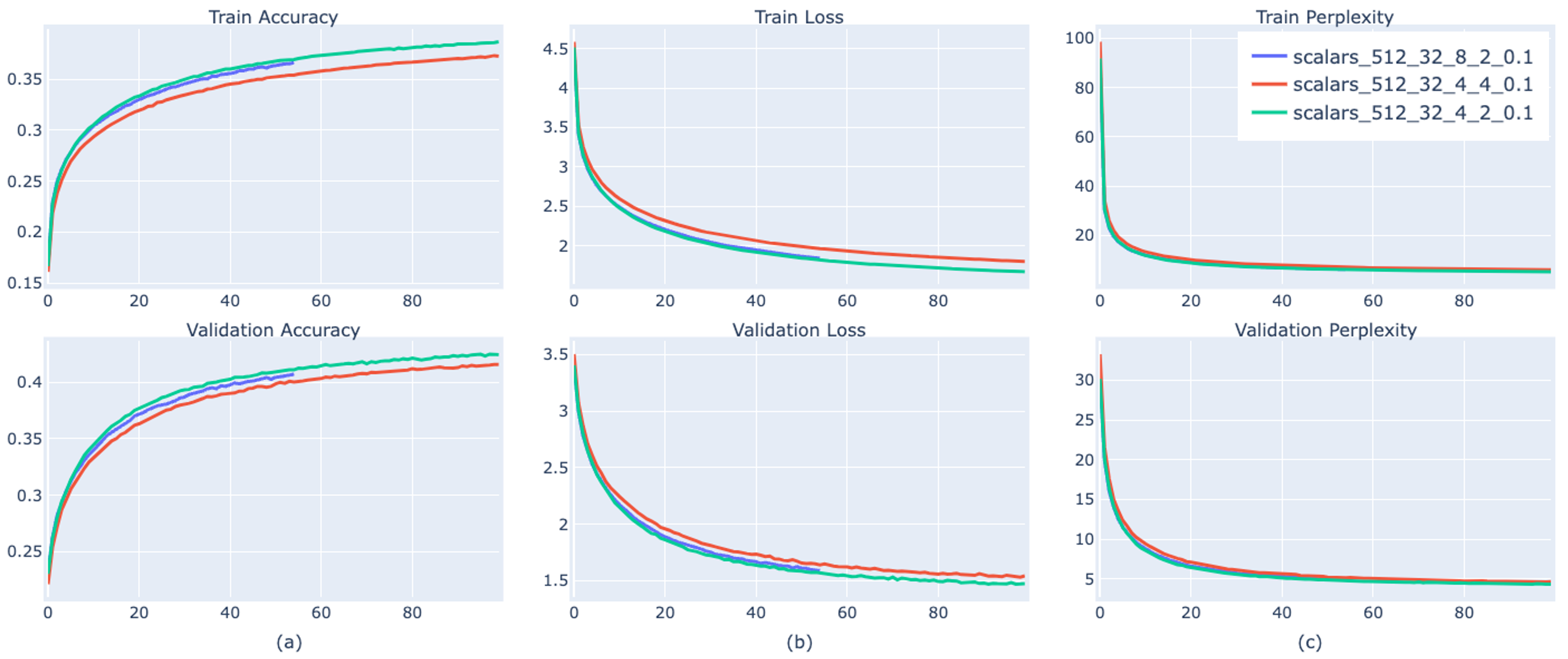}
\caption{Experiments: Maximum length = 512, Model Size = 32, number of heads = 4, 8, number of layers = 2, 4 and dropout sizes = 0.1 (combinations shown in legends). Ran for 100 epochs (x-axis). \textbf{(a)} Train and Validation Accuracy. \textbf{(b)} Train and Validation Loss. \textbf{(c)} Train and Validation Perplexity.}
\label{fig13}
\end{figure}

Next, we reduced the model size further to 32, keeping the layers at 2 and the dropout at 0.5, but varying the number of heads (4 and 8). With a model size of 32, 4 heads, 2 layers, and a dropout of 0.5, the model began overfitting within 100 epochs. When we increased the number of heads to 8 (for a model size of 32 and 2 layers), intriguingly, the model did not overfit. However, the validation accuracy was still higher, and the validation loss was lower than in the previous case. This could simply be attributed to the increase in heads, and it's possible that the model would have started overfitting after 100 epochs.
Next, we wanted to observe the effect of reducing the dropout from 0.5 to 0.1 (Figure \ref{fig13}). To our amazement, the model performed better in learning compared to the cases discussed in Figure \ref{fig12}. We tested a model size of 32, head sizes of 4 and 8, and a number of layers of 2 and 4, and found that the learning was comparable in all these cases. This suggested that with lower dropout, changes in the model head or layers might not significantly affect learning.
\begin{figure}[ht]
\centering
\includegraphics[width=150mm]{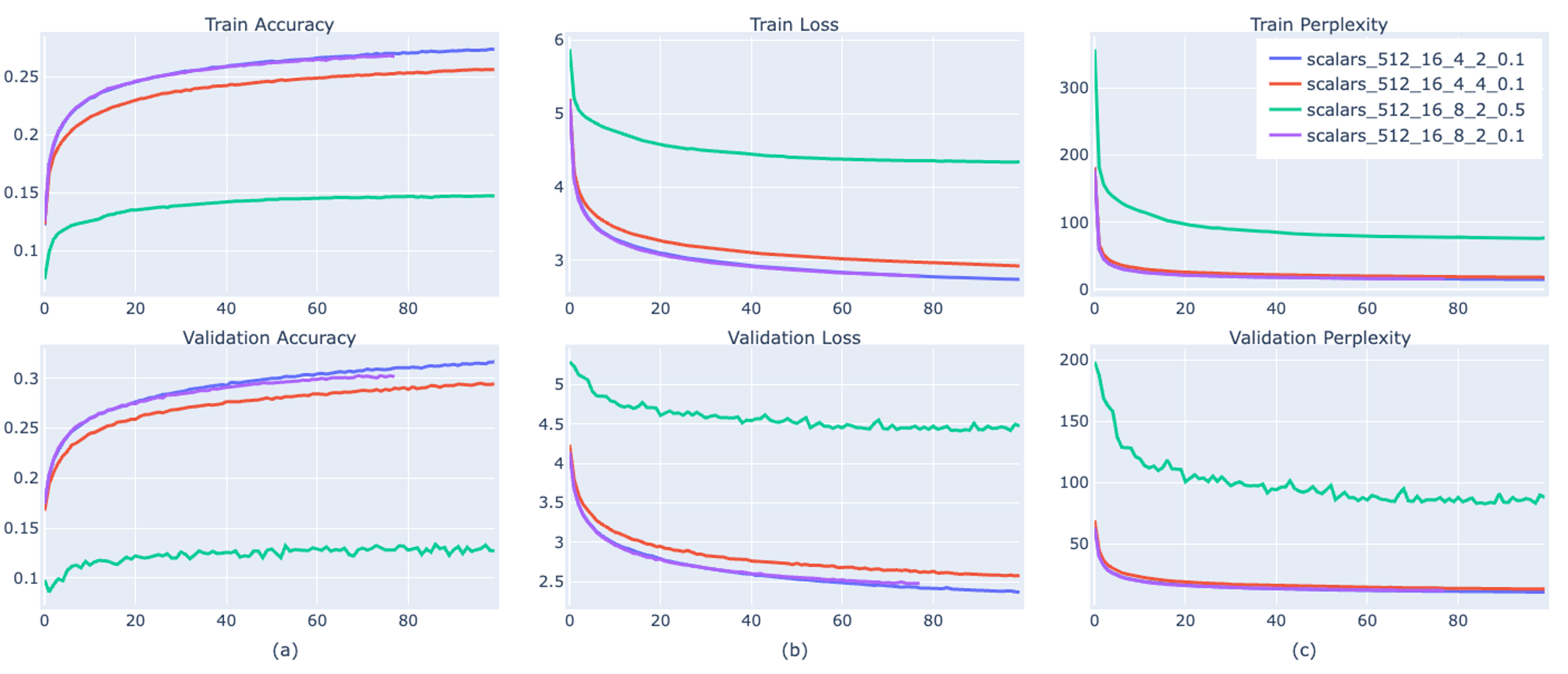}
\caption{Experiments: Maximum length = 512, Model Size = 16, number of heads = 4, 8, number of layers = 2, 4 and dropout sizes = 0.1, 0.5 (combinations shown in legends). Ran for 100 epochs (x-axis). \textbf{(a)} Train and Validation Accuracy. \textbf{(b)} Train and Validation Loss. \textbf{(c)} Train and Validation Perplexity.}
\label{fig14}
\end{figure}
\begin{figure}[ht]
\centering
\includegraphics[width=150mm]{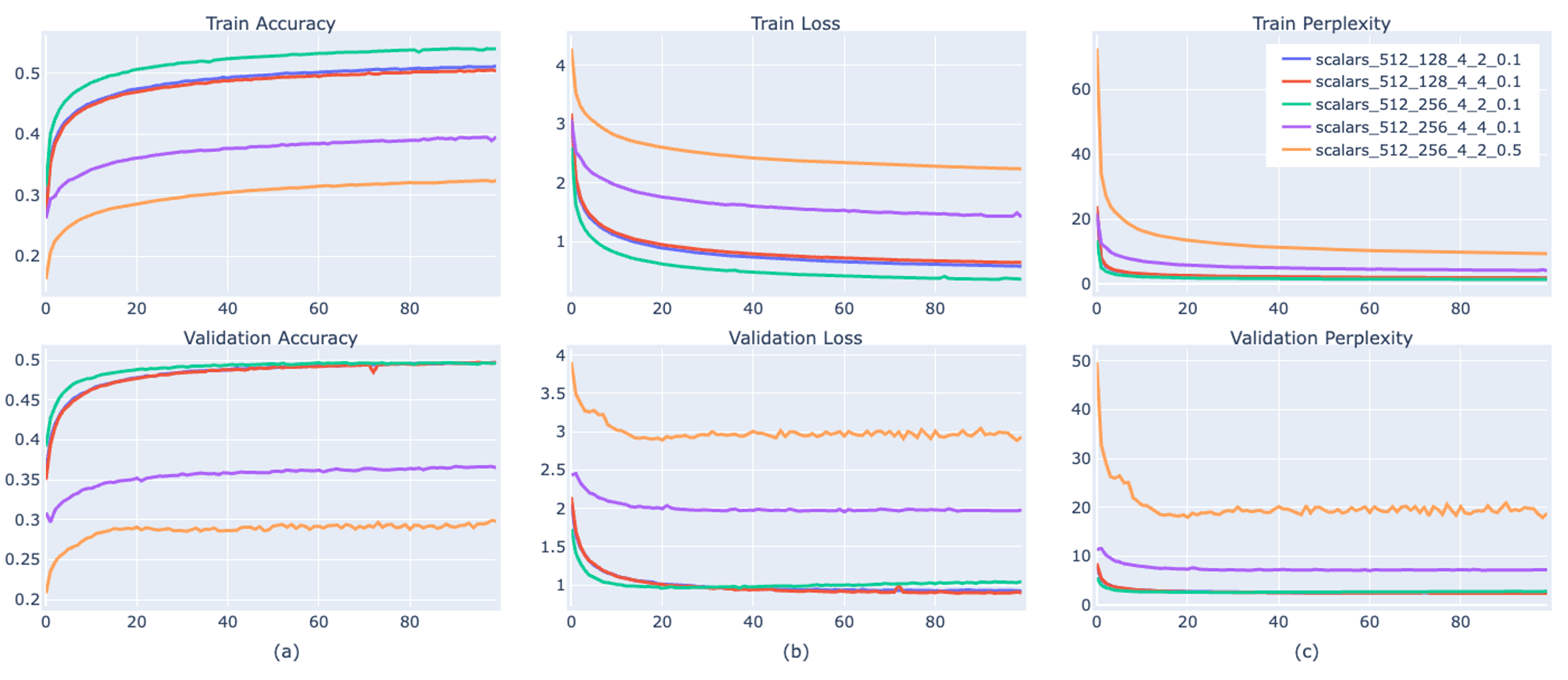}
\caption{Experiments: Maximum length = 512, Model Size = 128, 256, number of heads = 4, number of layers = 2, 4 and dropout sizes = 0.1, 0.5 (combinations shown in legends). Ran for 100 epochs (x-axis). \textbf{(a)} Train and Validation Accuracy. \textbf{(b)} Train and Validation Loss. \textbf{(c)} Train and Validation Perplexity.}
\label{fig15} 
\end{figure}
\begin{figure}[ht]
\centering
\includegraphics[width=150mm]{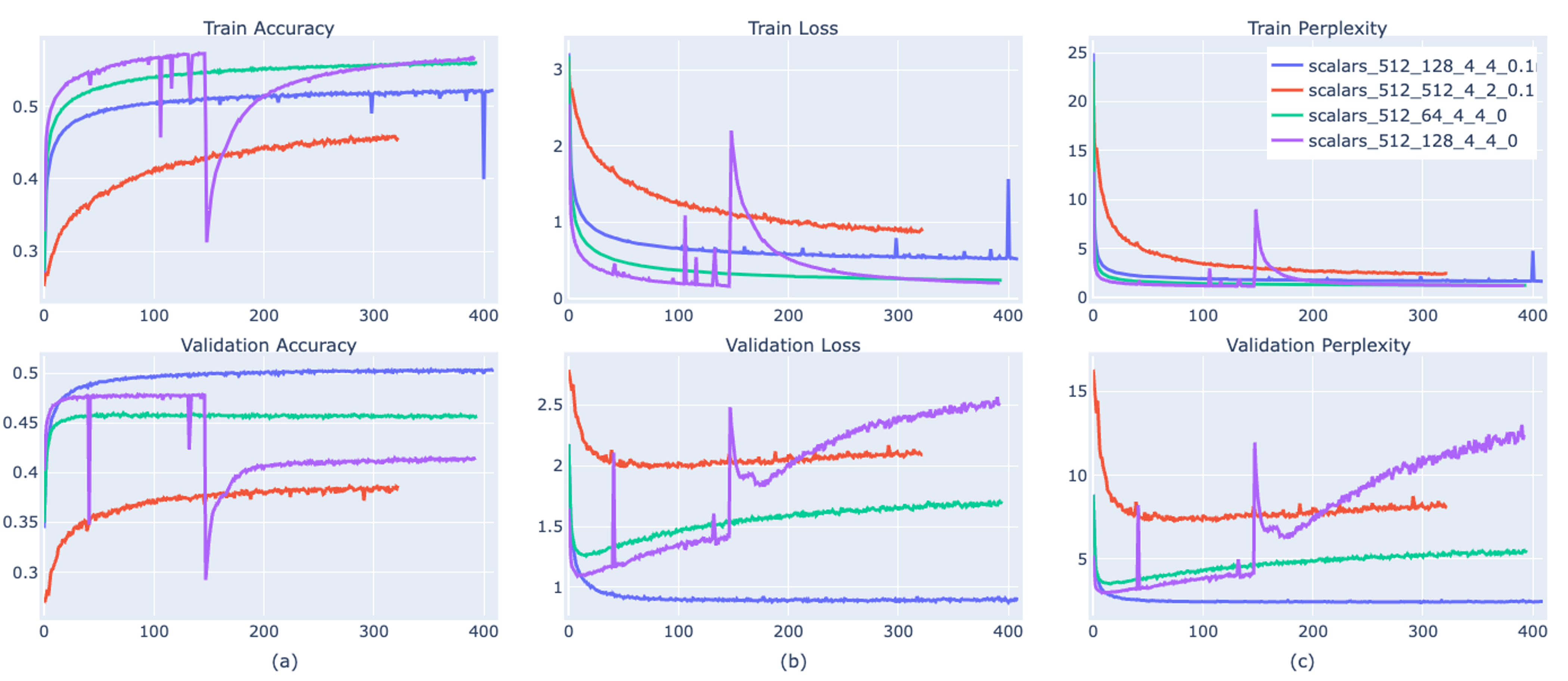}
\caption{Experiments: Maximum length = 512, Model Size = 64,128, 512, number of heads = 4, number of layers = 2, 4 and dropout sizes = 0.1, 0 (combinations shown in legends). Ran for 100 epochs (x-axis). \textbf{(a)} Train and Validation Accuracy. \textbf{(b)} Train and Validation Loss. \textbf{(c)} Train and Validation Perplexity.}
\label{fig16}
\end{figure}
\subsection{Results with the model size of 16}
In our subsequent experiment, we reduced the model size further to 16 (Figure \ref{fig14}), and again the model was tested with 4 and 8 heads, and 2 and 4 layers. In the case where the dropout was kept at 0.5, the learning was least effective, even with 8 heads for 2 layers. However, other cases with dropouts of 0.1 (and 4 or 8 heads with 2 layers) led us to rethink our previous approach of using higher dropouts like 0.5, 0.4, or 0.3. Overall, we observed that smaller dropout values yielded better results.
\subsection{Results with increasing reducing dropout value to 0.1}
Next, to test our hypothesis that a reduction in dropout value yields more stable results (i.e., from 0.5 to 0.1), we experimented with larger model sizes (128, 256) but kept the number of heads fixed at 4 and used only smaller encoder-decoder layers (2, 4). We observed that the model's accuracy was highest within 100 epochs for the configuration with a model size of 128, heads and layers equal to 4, and it was still learning even after 100 epochs. Increasing the model size to 256 reduced the validation accuracy and increased the loss, and changing the dropout from 0.1 to 0.5 further degraded learning. As evidenced by Table 1, it took just 26 million parameters (model size: 128, heads: 4, layers: 4) to achieve the best performance, and performance also degraded with an increase in the number of parameters.
\subsection{Best Results from Table \ref{table1}}
A comprehensive review of the table reveals an intriguing pattern that runs contrary to common expectations in model design; namely, that an increase in the number of parameters does not consistently correspond to a decrease in validation perplexity. This is vividly illustrated in the comparison between configurations with different model sizes, numbers of heads, layers, and dropout rates. Remarkably, the combination of a model size of 128, 4 heads, 4 layers, and a dropout rate of 0.1 achieved the best performance with just 26 million parameters. Even configurations with higher complexities and more parameters, such as those with 256 or 512 model sizes, failed to surpass this level of efficiency. Validation perplexity in some of these larger models escalated to alarming figures, as evidenced in specific rows of the table (Table \ref{table1})
\subsection{Investigating Best Results from Table \ref{table1} further by reducing dropout to 0 }
Lastly, we picked up the best configurations, according to Table 1, and we increased the number of epochs (Figure \ref{fig16}). This time we extended the runs between 350-400. In the best configuration (with model size of 128), we decrease the dropout to 0, to test the effect. We also reduced the model size to 64 and the dropout to 0, and we did one more test with the model size of 512. These tests were done to capture the effects on model performance with lower/zero dropouts, higher number of epochs, and range of parameters. We observed that decreasing the dropout to 0 in model size of 64 and 128 led to overfitting showing that for better performance dropouts are needed. Similarly, We observed that when the model size is bigger (model size of 512), it is still overfitted even with a dropout of 0.1. Perhaps, the higher dropout would have taken care of the overfitting, however, it's accuracy was way to less than the model size of 512. 
The investigation into various configurations of model sizes, numbers of heads, layers, and dropout values has unveiled a complex and multifaceted relationship between model complexity and learning efficacy. Contrary to conventional wisdom, the results consistently demonstrate that increased model complexity, as characterized by larger model sizes and more heads and layers, does not necessarily yield improved learning performance. In fact, certain simpler configurations, notably a model size of 128 with 4 heads and 4 layers and a lower dropout rate of 0.1, outperformed more complex structures, achieving superior performance with just 26 million parameters. This unexpected outcome emphasizes that a naive escalation in complexity and parameter count can not only inhibit learning but may lead to instability and overfitting. In a stark departure from the common pursuit of ever-larger models, these findings advocate for a more nuanced understanding of model architecture and hyperparameter tuning, balancing complexity against efficiency and stability.

\section{Conclusion}
\begin{itemize}
\item \textbf{Model Complexity vs. Efficiency}: The study's findings highlight that increasing model size and complexity does not necessarily lead to better learning performance. Moderate complexity in configurations at times outperformed larger, more intricate setups.
\item \textbf{Avoidance of Excessive Compute Power}: Rather than simply throwing more compute power and multiple GPUs at the problem, the study emphasizes a nuanced relationship between complexity and efficiency. It revealed that this approach could lead to instability, overfitting, and counterproductive results.
\item \textbf{Role of Dropout}: The analysis unveiled the multifaceted role of dropout in learning. The fine-tuning of this hyperparameter was shown to be vital, with too high or too low values leading to hindrances in convergence and learning.
\item \textbf{Overfitting Trends}: A balanced choice of model size, heads, layers, and dropout was found to be crucial for optimal learning without overfitting. Thoughtful hyperparameter tuning is preferred over a brute force increase in complexity.
\item \textbf{Influence of Heads and Layers}: These parameters had a nuanced impact on the learning process, illustrating the importance of understanding the complex interplay between different hyperparameters.
\item \textbf{Evaluation Metrics and Perplexity}: The study made significant use of validation loss curves and perplexity as indicators of learning stability and efficiency, focusing on these over traditional metrics like BLEU scores to understand overfitting and the impact of parameter increases on accuracy.
\item \textbf{Advocacy for Thoughtful Tuning}: Some simpler configurations achieved outstanding performance with only 26 million parameters. This demonstrates the importance of an intelligent, balanced approach that considers various hyperparameters instead of merely increasing resources and model sizes.
\item \textbf{Counterintuitive Findings}: The results challenged conventional wisdom, notably achieving the best performance with a configuration that was neither the largest nor most complex. It further emphasizes that optimization can often achieve better outcomes without relying on significant computational assets.
\item \textbf{Need for Hyperparameter Optimization}: Overall, the study underscores the importance of meticulous hyperparameter tuning. It acts as a reminder that less can be more if paired with an understanding of the intricacies of model architecture and careful optimization.
\end{itemize}


\bibliographystyle{plainnat}  
\bibliography{references}

\end{document}